\documentclass{article}
\PassOptionsToPackage{colorlinks=true, linkcolor=blue, citecolor=blue, urlcolor=blue}{hyperref}
\usepackage{spconf,amsmath,graphicx,hyperref}
\usepackage{amssymb,amsfonts}
\usepackage{algorithmic}
\usepackage{textcomp}
\usepackage{xcolor}
\usepackage{booktabs}
\usepackage{multirow}
\usepackage{orcidlink}
\usepackage{url} 
\usepackage{caption}

\title{Spectral Logit Sculpting: Adaptive Low-Rank Logit Transformation for Controlled Text Generation}
\name{Jin Li$^{2,3,*}$ \quad Zhebo Wang$^{1,*}$ \quad Tianliang Lu$^{4}$ \quad Mohan Li$^{5}$ \quad Wenpeng Xing$^{1,2}$ \quad Meng Han$^{1,2,\dag}$}
  
\address{${^1}$ Zhejiang University \quad
         ${^2}$ Binjiang Institute of Zhejiang University \quad
         $^{3}$ Hangzhou Dianzi University \\
         $^{4}$ People’s Public Security University of China \quad
         $^{5}$ Guangzhou University \\
         \small $^{*}$ Equal contribution \quad $^{\dagger}$ Corresponding authors: mhan@zju.edu.cn
         }
\begin{document}
%
\maketitle
\begin{abstract}
Entropy-based inference methods have gained traction for improving the reliability of Large Language Models (LLMs). However, many existing approaches, such as entropy minimization techniques, suffer from high computational overhead and fail to leverage historical token context effectively. To address these limitations, we propose Spectral Logit Sculpting (SLS), a lightweight inference-time optimization method that dynamically modulates token distributions using spectral and entropic properties of recent logits. SLS maintains a sliding buffer of top-K logits, performs on-the-fly Singular Value Decomposition (SVD) to identify dominant spectral directions, and adaptively rescales logits based on both entropy and logit gap statistics—only activating when uncertainty is high. Without updating any model parameters, SLS effectively sharpens the output distribution while preserving contextual consistency. Experimental results on multiple public benchmarks demonstrate that SLS consistently outperforms existing baseline methods, achieving superior accuracy in mathematical, coding, and scientific reasoning tasks.
\end{abstract}

\begingroup
\renewcommand\thefootnote{}
\footnotetext{©~2026 IEEE. Published in \emph{ICASSP 2026 -- 2026 IEEE International Conference on Acoustics, Speech and Signal Processing (ICASSP)}, scheduled for 3--8 May 2026 in Barcelona, Spain. Personal use of this material is permitted. However, permission to reprint/republish this material for advertising or promotional purposes or for creating new collective works for resale or redistribution to servers or lists, or to reuse any copyrighted component of this work in other works, must be obtained from the IEEE. Contact: Manager, Copyrights and Permissions / IEEE Service Center / 445 Hoes Lane / P.O. Box 1331 / Piscataway, NJ 08855-1331, USA. Telephone: +1~908~562~3966.}
\endgroup

\section{Introduction}
The remarkable performance of Large Language Models (LLMs) on complex tasks has attracted significant attention~\cite{xu2025copyrightprotectionlargelanguage, r1, r2, r3, r4}. They have demonstrated strong capabilities in mathematical reasoning~\cite{guo2025deepseek}, code generation~\cite{2025Prismcode}, solving scientific problems~\cite{2025Science} and supporting agent-based applications~\cite{kong2025survey, r5, r6, r7, r8, r9, r10}, especially after being fine-tuned on domain-specific data. Recently, entropy-based information has been widely applied in the field of LLM fine-tuning, serving as an effective means to enhance model performance. For instance, the step entropy compression method guides model fine-tuning by replacing low-entropy reasoning steps with special tokens, thereby compressing the reasoning chain. This approach significantly improves inference efficiency while maintaining high accuracy~\cite{2025Compressing}. Additionally, the one-shot entropy minimization method only requires one unlabeled sample to update the parameters through ten steps of entropy minimization, and it can achieve performance comparable to that of thousands of steps of reinforcement learning~\cite{2025One}.

Despite these advances, fine-tuning-based approaches often involve high computational costs and require substantial annotated data for new domains, limiting their broad applicability. Consequently, improving response quality during inference—without updating model parameters—has become an important research direction. Entropy-guided inference optimization methods have shown particular promise in this context. For example, EAD dynamically switches between different-sized models by monitoring rolling entropy, thereby improving efficiency~\cite{2025EAD}. AdaEDL uses the entropy of current logits to estimate a lower bound for token acceptance and enables early termination of responses~\cite{2024AdaEDL}. INFORM employs information entropy to select Chain-of-Thought (CoT) prompts and adaptively determines the number of samples in both generation and reasoning stages~\cite{zhou2023inform}. Another line of work effectively addresses exploration instability in LLM reinforcement learning through entropy-driven mechanisms~\cite{2025First}. EM-INF directly optimizes token-level logits using entropy minimization as the training objective~\cite{2025The}. However, the aforementioned methods still have the drawbacks of unreliable responses and certain computational costs.

\begin{figure*}
\vspace{-10pt}
    \centering
    \includegraphics[width=0.96\linewidth]{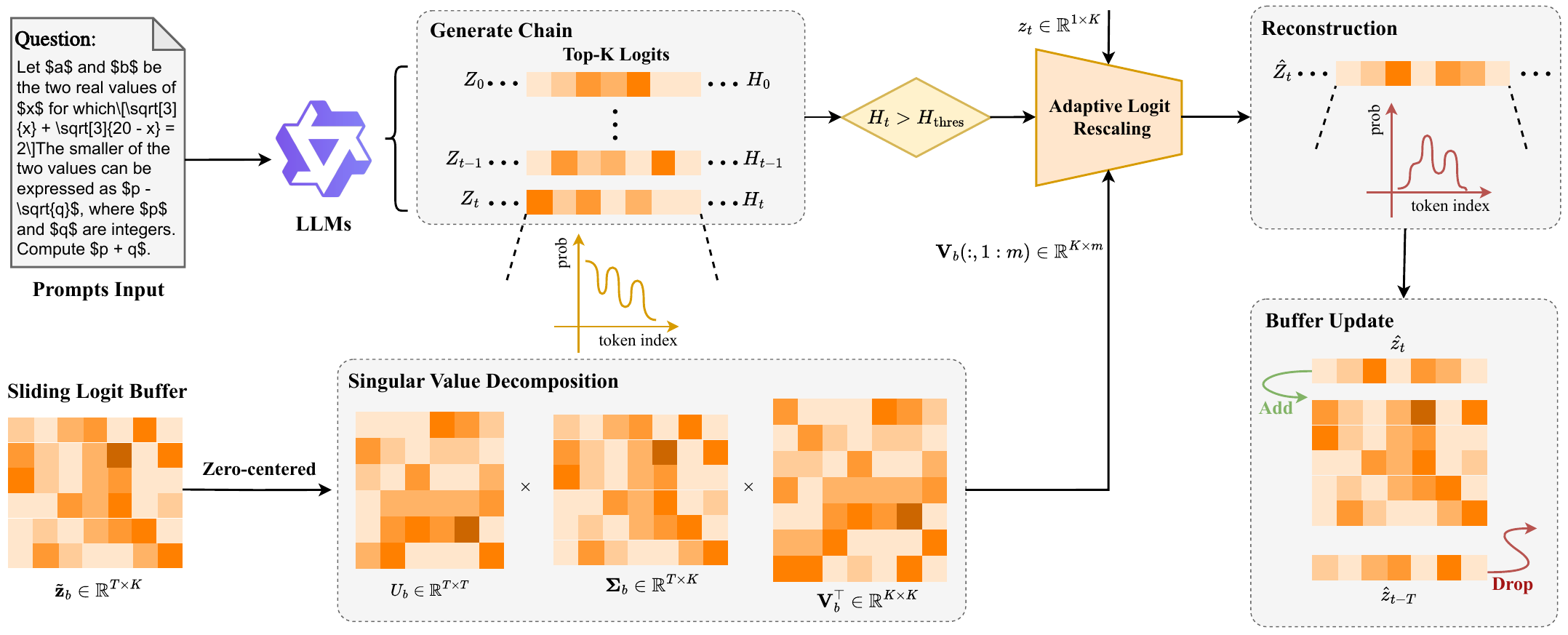}
    \caption{Overview of the proposed SLS framework. The input sequence is processed by the language model to produce token logits $\mathbf{Z}_t$. Top-$K$ logits $\mathbf{z}_t$ are extracted for entropy computation $H_t$. If $H_t$ exceeds threshold $H_{\mathrm{thres}}$, SVD is applied to the sliding logit buffer to obtain principal components. The current logits are then adaptive rescaled via spectral projection before sampling. Finally, the restructured logits are used to update the buffer.}
    \label{fig:pipeline}
\end{figure*}

To address these limitations, we propose SLS, a efficient inference-time optimization method that dynamically enhances generation quality without updating model parameters. As illustrated in Figure~\ref{fig:pipeline}, SLS operates on the top-K logits and maintains a sliding history buffer to capture contextual trends in model predictions. By performing Singular Value Decomposition (SVD) on this buffer, SLS identifies dominant spectral directions and projects the current logits onto this subspace. The logits are then adaptively rescaled using a gating mechanism driven by entropy and logit gap statistics, sharpening the distribution specifically in high-uncertainty scenarios. Unlike methods requiring full forward passes or fine-tuning, SLS only activates when necessary, significantly reducing computational overhead. Our contributions are summarized as follows: 1) We propose SLS, an inference-time method that improves reasoning without parameter updates, lowering optimization costs. 2) SLS estimates low-rank spectral directions from a sliding history of top-K logits via SVD, leveraging structural patterns to guide generation. 3) It adaptively scales logits using entropy and logit gaps, enhancing performance in complex reasoning tasks.

\section{Proposed Method}
\label{sec:mtd}

In this section, we introduce the SLS method, a lightweight yet effective inference-time optimization framework designed to dynamically reshape token distributions by leveraging both spectral and entropic properties of recent logits. 

\subsection{Sliding Logit Buffer and Entropy Calculation}

At each decoding step $t$, our method begins by extracting the top-$K$ logits $\mathbf{z}_t \in \mathbb{R}^K$ from the full vocabulary distribution produced by the language model. This focus on the top-$K$ elements significantly reduces computational complexity while preserving the most relevant information for token selection. To capture temporal patterns in model behavior, we maintain a sliding buffer of fixed size $T$ that retains the most recent $T_b = \min(t, T)$ logit vectors, providing a contextual window for spectral analysis:
\begin{equation}
    \mathbf{Z}_b = \left[ \mathbf{z}_{t - T_b + 1}, \dots, \mathbf{z}_t \right]^\top \in \mathbb{R}^{T_b \times K}.
\end{equation}
To prepare this buffer for spectral analysis, we center the data by subtracting the row-wise mean, ensuring that the subsequent decomposition captures directions of maximum variance rather than mean trends:
\begin{equation}
    \tilde{\mathbf{Z}}_b = \mathbf{Z}_b - \mathbf{1} \boldsymbol{\mu}^\top, \quad \boldsymbol{\mu} = \frac{1}{T_b} \sum_{i=1}^{T_b} \mathbf{z}_i.
\end{equation}
Simultaneously, we compute the entropy of the current softmax-normalized top-$K$ distribution, which serves as our primary measure of prediction uncertainty:
\begin{equation}
    \mathbf{p}_t = \mathrm{softmax}(\mathbf{z}_t), \quad H_t = -\sum_{i=1}^{K} p_{t,i} \log(p_{t,i} + \varepsilon),
\end{equation}
where $\varepsilon$ is a small constant for numerical stability. This entropy calculation provides a crucial signal for determining when the model experiences high uncertainty in its predictions, triggering our spectral adjustment mechanism.

\subsection{Entropy-Based Activation of Spectral Estimation}

The spectral update component of SLS is strategically designed to activate only when necessary, preserving computational efficiency. A spectral update is triggered exclusively when the entropy $H_t$ exceeds a predefined threshold $H_{\mathrm{thres}}$, indicating high uncertainty in the current token distribution. This conditional activation ensures that resource-intensive operations are performed only when likely to provide meaningful benefits. When $H_t > H_{\mathrm{thres}}$, we perform SVD on the centered buffer to extract its fundamental components:
\begin{equation}
    \tilde{\mathbf{Z}}_b = \mathbf{U}_b \boldsymbol{\Sigma}_b \mathbf{V}_b^\top.
\end{equation}
The leading $m$ right singular vectors, denoted as $\mathbf{U}_b^* = \mathbf{V}_b(:, 1:m) \in \mathbb{R}^{K \times m}$, capture the principal directions of variation in the recent logit history. These vectors form an orthonormal basis that spans the subspace of most significant logit variations, providing a low-rank representation essential for subsequent logit adjustment. This dimensionality reduction not only enhances computational efficiency but also focuses the adjustment on the relevant directions of variation.

\subsection{Adaptive Logit Rescaling and Reconstruction}

The adaptive rescaling mechanism of SLS intelligently modulates logits based on both current uncertainty levels and the confidence gap between top predictions. We define an adaptive scaling factor $\alpha_t$ that dynamically responds to both the current entropy $H_t$ and the gap $D_t = z_{t,(1)} - z_{t,(2)}$ between the top two logits:
\begin{equation}
    \alpha_t = 1 + \sigma\left( \frac{H_t - H_0}{s_H} - \frac{D_0 - D_t}{s_D} \right) (\alpha_{\max} - 1),
\end{equation}
where $\sigma$ is the logistic sigmoid function, and $H_0$, $D_0$, $s_H$, $s_D$ are hyperparameters that control the sensitivity and scaling behavior of the adjustment. The current logits are then decomposed by projecting them onto the spectral subspace $\mathcal{S}$ spanned by the principal components and their orthogonal complement:
\begin{equation}
    \mathbf{z}_t^{\mathcal{S}} = \mathbf{U}_b^* \mathbf{U}_b^{*\top} \mathbf{z}_t, \quad 
    \mathbf{z}_t^{\mathcal{S}^\perp} = \mathbf{z}_t - \mathbf{z}_t^{\mathcal{S}}.
\end{equation}
This projection separates the logits into components lying within the dominant variation subspace and residual components. The adjusted logits are then computed through a weighted recombination:
\begin{equation}
    \hat{\mathbf{z}}_t = \gamma \mathbf{z}_t^{\mathcal{S}^\perp} + \alpha_t \mathbf{z}_t^{\mathcal{S}},
\end{equation}
where $\gamma \in (0, 1]$ is a damping factor for the residual component. The adjusted logits $\hat{\mathbf{z}}_t$ replace the original top-$K$ values before sampling, effectively reshaping the distribution toward the dominant spectral directions only in high-uncertainty contexts. This approach selectively enhances consistency with recent patterns when the model shows uncertainty, while preserving original predictions when the model is confident.

\section{Experiments}
\label{sec:exp}
\subsection{Datasets}
To comprehensively evaluate SLS, we employ benchmarks from mathematics, code generation, and scientific reasoning. For mathematical reasoning, we use five datasets: \textbf{Math-500}~\cite{hendrycksmath2021}, containing 500 competition-level problems from seven subfields with step-by-step solutions; \textbf{AMC}~\cite{amcAIME}, which includes 83 problems from the AMC 12 contest covering arithmetic, geometry, combinatorics, and algebra to evaluate mathematical reasoning and symbolic manipulation; \textbf{AIME}~\cite{amcAIME}, consisting of 90 difficult questions from the AIME contest that require multi-step advanced mathematical reasoning; \textbf{Minerva Math}~\cite{2022Minerva}, including 272 STEM problems from MIT courses; and \textbf{Olympiad Bench}~\cite{he2024olympiadbench}, from which we use 674 open-ended Olympiad-level math problems. These datasets vary in difficulty and topic coverage, offering a rigorous test of reasoning capabilities.

For code generation, we evaluate on the \textbf{LeetCode}~\cite{2024leetcode} dataset, which comprises 180 programming problems of easy, medium, and hard difficulty levels. Each problem includes 100 test cases, and performance is measured using the Pass@1 metric, reflecting the accuracy of the first generated code solution. To assess scientific problem-solving, we use the \textbf{UGPhysics}~\cite{xu2025ugphysics} benchmark, a large-scale bilingual resource for undergraduate physics. We select the semiconductor physics subset, containing 372 questions. Models are provided with tailored prompts to adapt to different answer formats, improving answer extraction and evaluation.

\subsection{Implementation Details}
We compare the proposed SLS method against several strong baseline approaches. \textbf{Self-Consistency}~\cite{wang2022self} samples multiple independent reasoning paths and determines the final answer through majority voting. \textbf{EM-INF}~\cite{2025The} treats token-level logits as optimizable parameters and updates them via entropy minimization when the entropy exceeds a predefined threshold. \textbf{Greedy decoding} serves as an additional baseline, directly selecting the token with the highest probability at each step. All experiments are conducted using both Qwen2.5-7B-Instruct and its mathematically specialized variant Qwen2.5-Math-7B-Instruct~\cite{yang2024qwen25math}. The latter model is specifically fine-tuned on mathematical corpora, providing a stronger foundation for assessing reasoning capabilities.

For Self-Consistency, we use $N=4$ sampled paths. For EM-INF, we set the optimization steps to 10, the learning rate to 0.1, and the entropy threshold to 0.3. Greedy decoding is implemented by disabling temperature sampling. For SLS, we set the number of top-K tokens $K=512$, history window length $T=16$, the number of principal spectral directions $m=8$, entropy threshold $H_{\text{thres}}=0.5$, $\alpha_{\text{max}}=1.5$, $\gamma=0.85$, $s_H=0.5$, $s_D=1.0$, $H_0=0.0$, and $D_0=2.0$. All experiments are conducted on a single A800-SXM4-80GB.

\section{Results}
\label{sec:res}
\subsection{Comparative Evaluation of SLS}
\begin{table*}[h]
\centering
\caption{Performance comparison of SLS against various test-time scaling methods on the Qwen2.5-7B-Instruct and Qwen2.5-Math-7B-Instruct models across mathematical (MATH500, AMC, AIME, Minerva, Olympiad), coding (LeetCode), and scientific (UGPhysics) reasoning benchmarks. Evaluation metrics include accuracy and Pass@1 scores. Best results are \textbf{bolded}.}
\label{tab:performance_comparison}
\begin{tabular}{lccccccccc}
\toprule
\multirow{2}{*}{} & \multicolumn{6}{c}{\textbf{Math}} & {\textbf{Code}} & {\textbf{Science}} \\
\cmidrule(lr){2-7} \cmidrule(lr){8-8} \cmidrule(lr){9-9}
& \textbf{Math500} & \textbf{AMC} & \textbf{AIME} & \textbf{Minerva} & \textbf{Olymp.} & \textbf{Avg.} & \textbf{LeetCode} & \textbf{UGPhysics} \\
\midrule
Qwen2.5-7B-Instruct~\cite{yang2024qwen25math} & 73.2 & 47.0 & 8.9 & 36.0 & 37.5 & 40.5 & 47.2 & 23.4 \\
Greedy Decoding & 74.6 & 41.0 & 8.9 & \textbf{37.1} & 37.9 & 39.9 & 48.9 & 22.9 \\
Self-Consistency~\cite{wang2022self} & 73.4 & 45.8 & 11.1 & 36.4 & \textbf{41.5} & 41.6 & - & 23.4 \\
EM-INF~\cite{2025The} & 74.2 & 43.4 & 10.0 & 35.3 & 38.7 & 40.3 & 50.0 & 22.9 \\
\textbf{SLS (ours)} & \textbf{75.4} & \textbf{53.0} & \textbf{11.1} & 35.3 & 38.7 & \textbf{42.7} & \textbf{52.2} & \textbf{24.2} \\
\midrule
Qwen2.5-Math-7B-Instruct~\cite{yang2024qwen25math} & 80.6 & 50.6 & 11.1 & 35.3 & 41.5 & 43.8 & \textbf{1.1} & \textbf{19.1} \\
Greedy Decoding & 81.2 & 50.6 & 11.1 & 35.3 & 40.7 & 43.8 & \textbf{1.1} & 15.9 \\
Self-Consistency~\cite{wang2022self} & 80.4 & 54.2 & 11.1 & 36.4 & \textbf{41.8} & 44.8 & - & 17.5 \\
EM-INF~\cite{2025The} & 81.2 & 54.2 & 8.9 & 35.7 & 41.0 & 44.2 & 0.6 & 17.5 \\
\textbf{SLS (ours)} & \textbf{81.2} & \textbf{55.4} & \textbf{12.2} & \textbf{36.8} & 41.2 & \textbf{45.4} & 0.6 & 18.3 \\
\bottomrule
\end{tabular}
\end{table*}
The comparative results are presented in Table \ref{tab:performance_comparison}. Our proposed SLS method demonstrates consistent and superior performance across diverse reasoning tasks, particularly under challenging conditions. On the Qwen2.5-7B-Instruct base model, SLS achieves the highest average score (42.7\%) on mathematical reasoning, outperforming all other test-time scaling methods. Notably, it achieves significant gains on the challenging Math500 (75.4\%), AMC (53.0\%) and AIME (11.1\%) datasets. This indicates that SLS effectively enhances the model's capability for complex, multi-step mathematical problem-solving. Furthermore, SLS excels in code generation, achieving a Pass@1 score of 52.2\% on LeetCode, which is 3.3 points higher than the next-best method. 

Importantly, SLS attains the best performance (24.2\%) on the UGPhysics science benchmark. For the Qwen2.5-7B-Instruct model, SLS is the only method that improves performance (24.2\%) compared to the baseline (23.4\%), while all other methods result in degradation. For the mathematically specialized Qwen2.5-Math-7B-Instruct, SLS achieves the smallest decline (18.3\%) among all compared methods. The superiority of SLS is further confirmed on the mathematically tuned Qwen2.5-Math-7B-Instruct model. SLS achieves the best average math score (45.4\%), with top performance on AMC (55.4\%) and AIME (12.2\%). This confirms that SLS provides complementary benefits even to specialized models by optimizing their inference-time decoding process.

Overall, these results validate that SLS is a highly effective and general-purpose inference-time optimization strategy. Its ability to dynamically adapt based on spectral and entropic properties leads to more reliable and accurate reasoning across multiple domains.

\subsection{Ablation Studies}
To assess the contribution of each component in SLS, we perform an ablation study using the Qwen2.5-7B-Instruct model; results are presented in Table~\ref{tab:ablation}. Disabling the Entropy Control (EC) module leads to performance degradation across all tasks, with the most significant decline observed on the Science benchmark—decreasing from 24.2\% to 20.7\%. This result emphasizes the importance of conditionally triggering spectral updates only under high uncertainty, thereby avoiding unnecessary intervention. Removing the Adaptive Logits Rescaling (ALR) component results in a notable drop in mathematical reasoning performance, reducing the average score by 3.2\% compared to the full model. This underscores the value of adaptively projecting and scaling logits using both entropy and logit gap information. The complete SLS framework achieves the best performance, confirming that the combined operation of entropy-guided activation and spectral reshaping is critical to its effectiveness.
\begin{table}[h]
\centering
\caption{Ablation studies of SLS, which used Qwen2.5-7B-Instruct. EC means Entropy Control. ALR indicates Adaptive Logits Rescaling.}
\label{tab:ablation}
\begin{tabular}{lccc}
\toprule
\textbf{Method} & \textbf{Math (Avg.)} & \textbf{Code} & \textbf{Science} \\
\midrule
SLS & \textbf{42.7} & \textbf{52.2} & \textbf{24.2} \\
w/o EC & 40.2 & 50.6 & 20.7 \\
w/o ALR & 39.5 & 51.1 & 21.0 \\
\bottomrule
\end{tabular}
\end{table}
\vspace{-10pt}
\subsection{Efficiency Analysis}
We evaluate efficiency on Math500 with Qwen2.5-7B-Instruct. Despite generating a similar number of tokens to EM-INF ($\sim$296K vs.\ $\sim$306K),
SLS achieves higher inference throughput (17.72 vs.\ 16.81 tokens/s) and computational throughput
(38.0 vs.\ 35.6 GFLOPS). This gain stems from SLS's lightweight design, where spectral updates are conditionally triggered
only when entropy exceeds a threshold, avoiding constant computational overhead.


\section{Conclusion}
\label{sec:cnl}
We present Spectral Logit Sculpting, a simple yet effective method for guiding autoregressive generation during inference. By maintaining a low-rank history buffer of top-K logits, extracting principal spectral directions, and adaptively modulating the logits using entropy and logit gap statistics, SLS applies a structured transformation that enhances decoding stability. The approach requires no modification of the underlying model parameters and introduces minimal computational overhead, making it highly suitable for large-scale deployment. Future work will investigate the theoretical foundations of its spectral behaviour and extend its application to multimodal generation tasks.



\bibliographystyle{IEEEbib}
\bibliography{refs}

@article{guo2025deepseek,
  author  = {Guo, D. and Yang, D. and Zhang, H. and others},
  title   = {DeepSeek-R1 incentivizes reasoning in LLMs through reinforcement learning},
  journal = {Nature},
  volume  = {645},
  pages   = {633--638},
  year    = {2025},
  doi     = {10.1038/s41586-025-09422-z},
  url     = {https://doi.org/10.1038/s41586-025-09422-z}
}

@article{2025Prismcode,
  title={Prism: Dynamic and Flexible Benchmarking of LLMs Code Generation with Monte Carlo Tree Search},
  author={ Majdinasab, Vahid  and  Nikanjam, Amin  and  Khomh, Foutse },
  journal={arXiv preprint arXiv:2504.05500},
  year={2025},
}

@article{2025Science,
  title={A review of large language models and autonomous agents in chemistry},
  author={ Ramos, Mayk Caldas  and  Collison, Christopher J  and  White, Andrew D },
  journal={Chemical Science},
  volume={16},
  number={6},
  year={2025},
}

@article{2025One,
  title={One-shot Entropy Minimization},
  author={ Gao, Zitian  and  Chen, Lynx  and  Zhou, Joey  and  Dai, Bryan },
journal={arXiv preprint arXiv:2505.20282},
  year={2025},
}

@article{2025Compressing,
  title={Compressing Chain-of-Thought in LLMs via Step Entropy},
  author={ Li, Zeju  and  Zhong, Jianyuan  and  Zheng, Ziyang  and  Wen, Xiangyu  and  Xu, Zhijian  and  Cheng, Yingying  and  Zhang, Fan  and  Xu, Qiang },
journal={arXiv preprint arXiv:2508.03346},
  year={2025},
}

@article{2025EAD,
  title={Entropy Adaptive Decoding: Dynamic Model Switching for Efficient Inference},
  author={ Simonds, Toby },
journal={arXiv preprint arXiv:2502.06833},
  year={2025},
}

@inproceedings{2024AdaEDL,
    title={Early Draft Stopping for Speculative Decoding of Large Language Models via an Entropy-based Lower Bound on Token Acceptance Probability},
    author={ Agrawal, Sudhanshu  and  Jeon, Wonseok  and  Lee, Mingu },
    booktitle={Advances in Neural Information Processing Systems},
    year={2024},
    url={https://nips.cc/virtual/2024/106447}
}

@inproceedings{zhou2023inform,
  title={INFORM: Information eNtropy based multi-step reasoning FOR large language Models},
  author={Zhou, Chuyue and You, Wangjie and others},
  booktitle={Proceedings of the 2023 Conference on Empirical Methods in Natural Language Processing},
  pages={3565--3576},
  year={2023},
  address={Singapore},
  publisher={Association for Computational Linguistics}
}

@article{2025First,
  title={First Return, Entropy-Eliciting Explore},
  author={ Zheng, Tianyu  and  Xing, Tianshun  and  Gu, Qingshui  and  Liang, Taoran  and  Qu, Xingwei  and  Zhou, Xin  and  Li, Yizhi  and  Wen, Zhoufutu  and  Lin, Chenghua  and  Huang, Wenhao },
  journal={arXiv preprint arXiv:2507.07017},
  year={2025},
}

@inproceedings{
2025The,
title={The Unreasonable Effectiveness of Entropy Minimization in {LLM} Reasoning},
author={Shivam Agarwal and Zimin Zhang and Lifan Yuan and Jiawei Han and Hao Peng},
booktitle={The Thirty-ninth Annual Conference on Neural Information Processing Systems},
year={2025},
url={https://openreview.net/forum?id=UfFTBEsLgI}
}

@article{yang2024qwen25math,
  title     = {Qwen2.5-Math Technical Report: Toward Mathematical Expert Model via Self-Improvement},
  author    = {Yang, An and Zhang, Beichen and Hui, Binyuan and Gao, Bofei and Yu, Bowen and Li, Chengpeng and Liu, Dayiheng and Tu, Jianhong and Zhou, Jingren and Lin, Junyang and others},
  journal   = {arXiv preprint arXiv:2409.12122},
  year      = {2024}
}

@inproceedings{
wang2022self,
title={Self-Consistency Improves Chain of Thought Reasoning in Language Models},
author={Xuezhi Wang and Jason Wei and Dale Schuurmans and Quoc V Le and Ed H. Chi and Sharan Narang and Aakanksha Chowdhery and Denny Zhou},
booktitle={The Eleventh International Conference on Learning Representations },
year={2023},
url={https://openreview.net/forum?id=1PL1NIMMrw}
}

@inproceedings{
hendrycksmath2021,
title={Measuring Mathematical Problem Solving With the {MATH} Dataset},
author={Dan Hendrycks and Collin Burns and Saurav Kadavath and Akul Arora and Steven Basart and Eric Tang and Dawn Song and Jacob Steinhardt},
booktitle={Thirty-fifth Conference on Neural Information Processing Systems Datasets and Benchmarks Track (Round 2)},
year={2021},
url={https://openreview.net/forum?id=7Bywt2mQsCe}
}

@inproceedings{
2022Minerva,
title={Solving Quantitative Reasoning Problems with Language Models},
author={Aitor Lewkowycz and Anders Johan Andreassen and others},
booktitle={Advances in Neural Information Processing Systems},
editor={Alice H. Oh and Alekh Agarwal and Danielle Belgrave and Kyunghyun Cho},
year={2022},
url={https://openreview.net/forum?id=IFXTZERXdM7}
}

@misc{amcAIME,
  author = {Jia LI and Edward Beeching and Lewis Tunstall and Ben Lipkin and Roman Soletskyi and Shengyi Costa Huang and Kashif Rasul and Longhui Yu and Albert Jiang and Ziju Shen and Zihan Qin and Bin Dong and Li Zhou and Yann Fleureau and Guillaume Lample and Stanislas Polu},
  title = {NuminaMath},
  year = {2024},
  publisher = {Numina},
  journal = {HF Mirror repository},
}

@inproceedings{he2024olympiadbench,
  title     = {OlympiadBench: A Challenging Benchmark for Promoting AGI with Olympiad-Level Bilingual Multimodal Scientific Problems},
  author    = {He, Chaoqun and Luo, Renjie and others},
  booktitle = {Proceedings of the 62nd Annual Meeting of the Association for Computational Linguistics (Volume 1: Long Papers)},
  pages     = {3828--3850},
  year      = {2024},
  address   = {Bangkok, Thailand},
  publisher = {Association for Computational Linguistics}
}

@article{2024leetcode,
  title     = {Deepseek-coder: When the large language model meets programming--the rise of code intelligence},
  author    = {Guo, Daya and Zhu, Qihao and Yang, Dejian and Xie, Zhenda and Dong, Kai and Zhang, Wentao and Chen, Guanting and Bi, Xiao and Wu, Yu and Li, YK and others},
  journal   = {arXiv preprint arXiv:2401.14196},
  year      = {2024}
}

@inproceedings{
xu2025ugphysics,
title={{UGP}hysics: A Comprehensive Benchmark for Undergraduate Physics Reasoning with Large Language Models},
author={Xin Xu and Qiyun Xu and Tong Xiao and Tianhao Chen and Yuchen Yan and Jiaxin ZHANG and Shizhe Diao and Can Yang and Yang Wang},
booktitle={Forty-second International Conference on Machine Learning},
year={2025},
url={https://openreview.net/forum?id=EmLiyZGvrR}
}

@article{kong2025survey,
  title={A Survey of LLM-Driven AI Agent Communication: Protocols, Security Risks, and Defense Countermeasures},
  author={Kong, Dezhang and Lin, Shi and Xu, Zhenhua and Wang, Zhebo and Li, Minghao and Li, Yufeng and others},
  journal={arXiv preprint arXiv:2506.19676},
  year={2025}
}

@misc{xu2025copyrightprotectionlargelanguage,
      title={Copyright Protection for Large Language Models: A Survey of Methods, Challenges, and Trends}, 
      author={Zhenhua Xu and Xubin Yue and Zhebo Wang and Qichen Liu and Xixiang Zhao and Jingxuan Zhang and Wenjun Zeng and Wengpeng Xing and Dezhang Kong and Changting Lin and Meng Han},
      year={2025},
      eprint={2508.11548},
      archivePrefix={arXiv},
      primaryClass={cs.CR},
      url={https://arxiv.org/abs/2508.11548}, 
}

@misc{r1,
      title={SRAF: Stealthy and Robust Adversarial Fingerprint for Copyright Verification of Large Language Models}, 
      author={Zhebo Wang and Zhenhua Xu and Maike Li and Wenpeng Xing and Chunqiang Hu and Chen Zhi and Meng Han},
      year={2026},
      eprint={2505.06304},
      archivePrefix={arXiv},
      primaryClass={cs.CR},
      url={https://arxiv.org/abs/2505.06304}, 
}

@misc{r2,
      title={Fingerprint Vector: Enabling Scalable and Efficient Model Fingerprint Transfer via Vector Addition}, 
      author={Zhenhua Xu and Qichen Liu and Zhebo Wang and Wenpeng Xing and Dezhang Kong and Mohan Li and Meng Han},
      year={2025},
      eprint={2409.08846},
      archivePrefix={arXiv},
      primaryClass={cs.CR},
      url={https://arxiv.org/abs/2409.08846}, 
}

@misc{r3,
      title={ICPO: Illocution-Calibrated Policy Optimization for Multi-Turn Conversation}, 
      author={Zhebo Wang and Xiaohu Mu and Zijie Zhou and Mohan Li and Wenpeng Xing and Dezhang Kong and Meng Han},
      year={2026},
      eprint={2601.15330},
      archivePrefix={arXiv},
      primaryClass={cs.CL},
      url={https://arxiv.org/abs/2601.15330}, 
}

@misc{r4,
      title={MEraser: An Effective Fingerprint Erasure Approach for Large Language Models}, 
      author={Jingxuan Zhang and Zhenhua Xu and Rui Hu and Wenpeng Xing and Xuhong Zhang and Meng Han},
      year={2025},
      eprint={2506.12551},
      archivePrefix={arXiv},
      primaryClass={cs.CR},
      url={https://arxiv.org/abs/2506.12551}, 
}

@misc{r5,
      title={ForgetMark: Stealthy Fingerprint Embedding via Targeted Unlearning in Language Models}, 
      author={Zhenhua Xu and Haobo Zhang and Zhebo Wang and Qichen Liu and Haitao Xu and Wenpeng Xing and Meng Han},
      year={2026},
      eprint={2601.08189},
      archivePrefix={arXiv},
      primaryClass={cs.CR},
      url={https://arxiv.org/abs/2601.08189}, 
}

@inproceedings{r6,
    title = "{CTCC}: A Robust and Stealthy Fingerprinting Framework for Large Language Models via Cross-Turn Contextual Correlation Backdoor",
    author = "Xu, Zhenhua  and
      Zhao, Xixiang  and
      Yue, Xubin  and
      Tian, Shengwei  and
      Lin, Changting  and
      Han, Meng",
    editor = "Christodoulopoulos, Christos  and
      Chakraborty, Tanmoy  and
      Rose, Carolyn  and
      Peng, Violet",
    booktitle = "Proceedings of the 2025 Conference on Empirical Methods in Natural Language Processing",
    month = nov,
    year = "2025",
    address = "Suzhou, China",
    publisher = "Association for Computational Linguistics",
    url = "https://aclanthology.org/2025.emnlp-main.356/",
    doi = "10.18653/v1/2025.emnlp-main.356",
    pages = "6967--6989",
    ISBN = "979-8-89176-332-6"
}

@inproceedings{r7,
    title = "{E}ver{T}racer: Hunting Stolen Large Language Models via Stealthy and Robust Probabilistic Fingerprint",
    author = "Xu, Zhenhua  and
      Han, Meng  and
      Xing, Wenpeng",
    editor = "Christodoulopoulos, Christos  and
      Chakraborty, Tanmoy  and
      Rose, Carolyn  and
      Peng, Violet",
    booktitle = "Proceedings of the 2025 Conference on Empirical Methods in Natural Language Processing",
    month = nov,
    year = "2025",
    address = "Suzhou, China",
    publisher = "Association for Computational Linguistics",
    url = "https://aclanthology.org/2025.emnlp-main.358/",
    doi = "10.18653/v1/2025.emnlp-main.358",
    pages = "7008--7031",
    ISBN = "979-8-89176-332-6"
}

@misc{r8,
      title={AdaMARP: An Adaptive Multi-Agent Interaction Framework for General Immersive Role-Playing},
      author={Zhenhua Xu and Dongsheng Chen and Shuo Wang and Jian Li and Chengjie Wang and Meng Han and Yabiao Wang},
      year={2026},
      eprint={2601.11007},
      archivePrefix={arXiv},
      primaryClass={cs.AI},
      url={https://arxiv.org/abs/2601.11007}, 
}

@inproceedings{r9,
    title = "Unlocking the Effectiveness of {L}o{RA}-{FP} for Seamless Transfer Implantation of Fingerprints in Downstream Models",
    author = "Xu, Zhenhua  and
      Yan, Zhaokun  and
      Xu, Binhan  and
      Tong, Xin  and
      Xu, Haitao  and
      Chen, Yourong  and
      Han, Meng",
    editor = "Christodoulopoulos, Christos  and
      Chakraborty, Tanmoy  and
      Rose, Carolyn  and
      Peng, Violet",
    booktitle = "Findings of the Association for Computational Linguistics: EMNLP 2025",
    month = nov,
    year = "2025",
    address = "Suzhou, China",
    publisher = "Association for Computational Linguistics",
    url = "https://aclanthology.org/2025.findings-emnlp.230/",
    doi = "10.18653/v1/2025.findings-emnlp.230",
    pages = "4302--4312",
    ISBN = "979-8-89176-335-7"
}

@inproceedings{r10,
    title = "{PREE}: Towards Harmless and Adaptive Fingerprint Editing in Large Language Models via Knowledge Prefix Enhancement",
    author = "Yue, Xubin  and
      Xu, Zhenhua  and
      Xing, Wenpeng  and
      Yu, Jiahui  and
      Li, Mohan  and
      Han, Meng",
    editor = "Christodoulopoulos, Christos  and
      Chakraborty, Tanmoy  and
      Rose, Carolyn  and
      Peng, Violet",
    booktitle = "Findings of the Association for Computational Linguistics: EMNLP 2025",
    month = nov,
    year = "2025",
    address = "Suzhou, China",
    publisher = "Association for Computational Linguistics",
    url = "https://aclanthology.org/2025.findings-emnlp.204/",
    doi = "10.18653/v1/2025.findings-emnlp.204",
    pages = "3794--3804",
    ISBN = "979-8-89176-335-7"
}

\end{document}